\documentclass[letterpaper, 10 pt, conference]{ieeeconf}  % Comment this line out if you need a4paper

\IEEEoverridecommandlockouts                              % This command is only needed if 
\usepackage{graphicx}
\graphicspath{{figures/}}
\usepackage{xcolor}
\usepackage[colorlinks]{hyperref}
\title{\LARGE \bf
Challenges in Close-Proximity Safe and Seamless Operation of Manned
and Unmanned Aircraft in Shared Airspace}

\author{Jay Patrikar$^{1}$, Joao P. A. Dantas$^{1,2}$, Sourish Ghosh$^{1}$, Parv Kapoor$^{1}$, Ian Higgins$^{1}$,\\ Jasmine J. Aloor$^{3}$, Ingrid Navarro$^{1}$, Jimin Sun$^{1}$, Ben Stoler$^{1}$, Milad Hamidi$^{1}$,\\ Rohan Baijal$^{4}$, Brady Moon$^{1}$, Jean Oh$^{1}$, Sebastian Scherer$^{1}$% <-this % stops a space
\thanks{$^{1}$Authors are with the Robotics Institute, Carnegie Mellon University, Pittsburgh, PA, USA. {\tt\small \{jpatrika, jdantas, sourishg, parvk, ihiggins, ingridn, jimins2, bstoler, mmoghass, bradym, hyaejino, basti\}@andrew.cmu.edu}}
\thanks{$^{2}$Joao P. A. Dantas is with the Decision Support Systems Subdivision, Institute for Advanced Studies, Sao Jose dos Campos, SP, Brazil. {\tt\small dantasjpad@fab.mil.br}}
\thanks{$^{3}$Jasmine J. Aloor is with the Indian Institute of Technology, Kharagpur, WB, India. {\tt\small jasminejerry@iitkgp.ac.in}}
\thanks{$^{4}$Rohan Baijal is with the Indian Institute of Technology, Kanpur, UP, India. {\tt\small rbaijal@iitk.ac.in}}
}

\begin{document}

\maketitle
\thispagestyle{empty}
\pagestyle{empty}

%%%%%%%%%%%%%%%%%%%%%%%%%%%%%%%%%%%%%%%%%%%%%%%%%%%%%%%%%%%%%%%%%%%%%%%%%%%%%%%%
\begin{abstract}

We propose developing an integrated system to keep autonomous unmanned aircraft safely separated and behave as expected in conjunction with manned traffic. The main goal is to achieve safe manned-unmanned vehicle teaming to improve system performance, have each (robot/human) teammate learn from each other in various aircraft operations, and reduce the manning needs of manned aircraft. The proposed system anticipates and reacts to other aircraft using natural language instructions and can serve as a co-pilot or operate entirely autonomously. We point out the main technical challenges where improvements on current state-of-the-art are needed to enable Visual Flight Rules to fully autonomous aerial operations, bringing insights to these critical areas. Furthermore, we present an interactive demonstration in a prototypical scenario with one AI pilot and one human pilot sharing the same terminal airspace, interacting with each other using language, and landing safely on the same runway. We also show a demonstration of a vision-only aircraft detection system.  \href{https://youtu.be/iU_MyMwuE8E}{[Video]}\footnote{ \href{https://youtu.be/iU_MyMwuE8E}{Video: https://youtu.be/iU\_MyMwuE8E}}  

\end{abstract}

%%%%%%%%%%%%%%%%%%%%%%%%%%%%%%%%%%%%%%%%%%%%%%%%%%%%%%%%%%%%%%%%%%%%%%%%%%%%%%%%
\section{Introduction}
\label{sec:1}

 Advanced Aerial Mobility (AAM) is an inclusive term that covers urban (UAM), regional (RAM), intraregional (IAM), and suburban air mobility (SAM) solutions~ \cite{aweiss2018}. All of these proposed solutions have one thing in common: They all envision a future where autonomous or semi-autonomous aerial vehicles are seamlessly integrated into the current airspace system. AAM solutions open doors to significant socio-economic benefits while at the same time presenting challenges in the seamless integration of these systems with human-operated aircraft and controlling agencies.

Today, manned and unmanned vehicles are separated, limiting the utility and flexibility of operations and reducing efficiency. Human-operated aircraft follow one of the two rules for operation: Visual Flight Rules (VFR) or Instrument Flight Rules (IFR). Flying under VFR or IFR is typically a function of weather conditions. Under VFR, an aircraft is flown just like driving a car within the Federal Aviation Administration (FAA) established rules of the road. On the other hand, IFR flying is typically associated with flying aircraft under degraded weather conditions where separation is provided by the Air Traffic Control (ATC). While the pursuit of integrating AAM can start either under IFR or VFR, automated-VFR operations often have better scalability than automated-IFR operations~\cite{mueller2017enabling}. Another option involves Unmanned Aircraft System Traffic Management (UTM) solution~\cite{aweiss2018}, which in its current iteration only focuses on small unmanned aerial aircraft operating close to the ground (less than 700 ft) in uncontrolled airspace.

Mastering VFR operations for autonomous aircraft has significant operational advantages at unimproved sites and achievable traffic density compared to IFR or completely separated operations between manned and unmanned systems. For (semi-)autonomous aircraft to operate in tandem with human pilots and ATC controllers under VFR, technological advancements in certain key areas are required, specifically:

\begin{enumerate}
\item Unmanned aircraft should be able to guarantee self-separation even in a tightly-spaced terminal airspace environment;
\item Unmanned vehicles should be able to interpret high-level instructions by ATC to meet the expectations of a normal traffic flow;
\item Autonomous aircraft need to understand the expected and unexpected behavior of other aircraft;
\item Communications by other pilots and ATC need to be parsed, and corresponding responses should be produced; and
\item Other aircraft need to be detected and estimated, and their future trajectories need to be predicted.
\end{enumerate}

Many of these challenges have parallels in the self-driving industry, and technological improvements can be leveraged to produce domain-specific solutions for AAM. While this is promising, VFR-like AAM integration introduces newer challenges while pushing boundaries on the current state of technology.

% Summary
% The rest of this paper is organized as follows. Section~\ref{sec:2} describes the technical challenges pointed out to solve the problem statement. Section~\ref{sec:2} brings an interactive demonstration of the problem statement. Finally, in Section~\ref{sec:4} we have the conclusions about the theme exposed in this work.

\section{Technical Challenges}
\label{sec:2}

% This section brings the critical areas of development where progress on current state-of-the-art are required to promote VFR-like autonomous aerial operations.

This section brings the key areas of development where improvements on current state-of-the-art are needed to enable VFR-like autonomous aerial operations. Also, we bring some solutions considering the technical challenges pointed out.

\subsection{Aircraft Detection and Tracking}

\begin{figure}[thpb]
\centering
\includegraphics[width=0.485\textwidth]{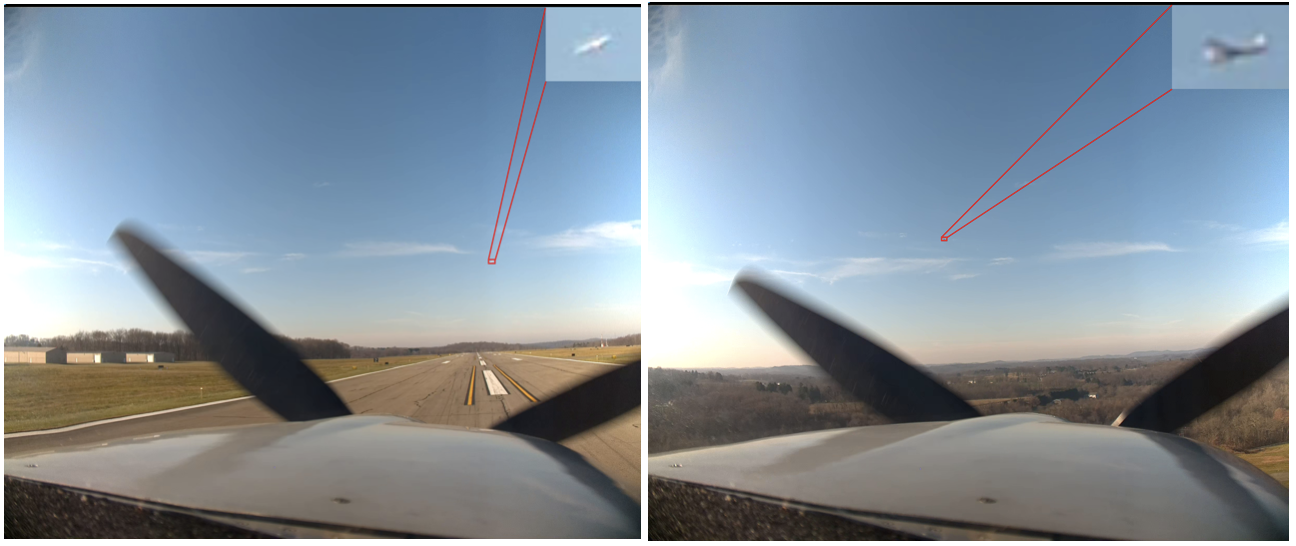}
\caption{Visualization of the detected aircraft bounding boxes from the data onboard a general aviation aircraft. SnR is really poor at long ranges, and our algorithms are trained to detect aircraft even in low SnR situations.}
\label{fig:1}
\end{figure}

See-and-Avoid is one of the key tenets of VFR operation. The ability to spot other aircraft or aerial hazards like birds, balloons, gliders, etc., and execute maneuvers to mitigate a collision hazard is critical to successfully deploying AAM solutions in the National Airspace System (NAS). Detecting aircraft at long ranges comes with a lot of challenges. Some of these are outlined as follows:

\begin{itemize}

\item \textbf{Low SnR}: Aircraft or other flying objects at long ranges appear very small and typically have very low signal-to-noise (SnR) ratios (see Fig.~\ref{fig:1})

\item \textbf{Poor performance on small objects}: The state-of-the-art in object detection keeps changing frequently, and the standard benchmarks such as COCO~\cite{lin2014microsoft} are constantly being beaten from time to time as more complex models are created. However, the performance of these detectors on small objects ($<32^{2}$ px$^{2}$) is pretty poor as they have low average precision and recall. In the context of our work, the performance is even worse since most of our data consist of small objects.
\item \textbf{Computational constraints}: The SOTA detectors are typically very complex regarding the number of learnable parameters, and they usually work on low to moderate image resolutions. Since our context is primarily small objects, we need to operate on high-res 2K images. Such images cannot be processed without downsampling by modern detectors.
\end{itemize}

\subsection{Intent Prediction}

Reasoning about the potential future trajectories that other aircraft can take is critical to ensure that the best actions are taken that minimize risk. Typical prediction methods use short horizons and fail to capture the long-term intent of other aircraft. Most trajectory prediction work has been explored in the pedestrian and autonomous vehicle domains. Within AAM, goal points such as airports, pilots, ATC communications, and rules of way (such as FAR §91.113) are often known. Using this explicit source of information and implicit sources like weather can help decipher the intent of other aircraft and increase the length of reliable predictions.

\begin{figure*}[thpb]
\centering
\includegraphics[width=\textwidth]{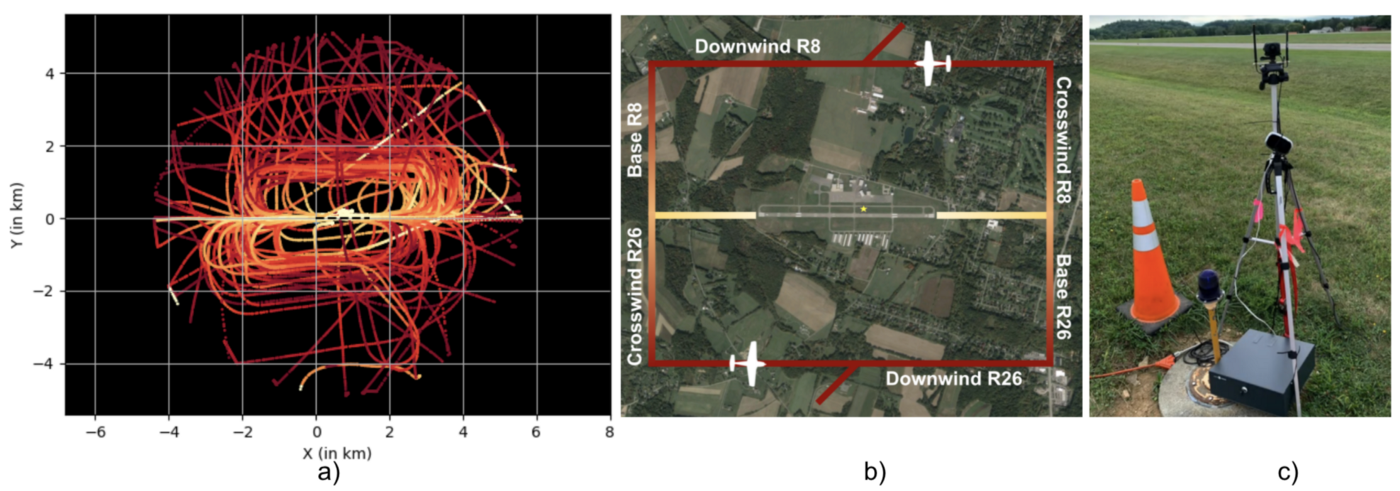}
\caption{The figure \cite{patrikar2021trajair} shows the dataset and its collection setup at the Pittsburgh-Butler Regional Airport (KBTP) — a non-towered GA airport that serves as a primary location for the dataset. Lighter color indicates a lower altitude. a) Shown is a snippet of the processed dataset with aircraft trajectories showing clear lobes for traffic patterns for both runways. b) The left traffic pattern and terminology for the runways at the airport. c) Picture of the data-collection setup.}
\label{fig:2}
\end{figure*}

The lack of datasets and baseline methods makes it difficult to conduct meaningful research. Toward this end, we have collected traffic data from transponders at select controlled and uncontrolled airports to understand and train different models for intent prediction. To capture the influence of weather on pilot decisions, the weather data was also collected by parsing Meteorological Aerodrome Reports (METARs) to gather the relevant sections from the full weather report. We recently published the first chunk of the trajectory data TrajAir along with a baseline method christened TrajAirNet to predict aircraft trajectories trained on the collected data (see Fig.~\ref{fig:2})~\cite{patrikar2021trajair}.

\subsection{Safe and Socially-Compliant Navigation}

\begin{figure*}[thpb]
\centering
\includegraphics[width=\textwidth]{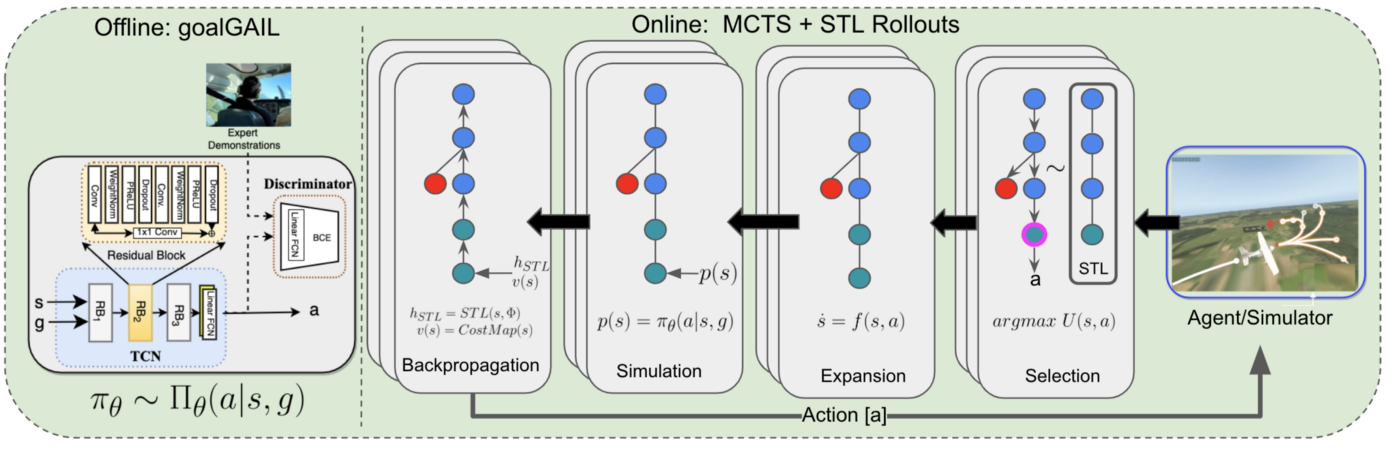}
\caption{Figure shows the proposed planning setup for a single agent. The offline policy is trained using a Generative Adversarial Imitation Learning algorithm that learns system behavior from the recorded trajectories from KBTP airport. Online, this behavior is executed using a Monte Carlo Tree Search algorithm that uses multiple roll-outs to model possible future states of the human agent. The roll-outs are constrained using Signal Temporal Logic specifications that ensure that the actions follow the rules established by the FAA.}
\label{fig:3}
\end{figure*}

As the FAR rules only specify a rough guideline, autonomous vehicles must be able to make flexible decisions to comply with the traffic norm of arbitrary situations. The idea of social navigation is to learn to follow the observed traffic patterns in a socially-compliant manner. Generating actions that are not only safe but also socially compliant while following rules is thus critical in developing behavior that is acceptable to human pilots co-habiting the same airspace. We use variants of Monte Carlo Tree Search algorithms that combine learned behavior with logic specifications like Signal Temporal Logic to enable safe and socially compliant navigation (see Fig.~\ref{fig:3}).

\subsection{Guaranteeing safety}

The safety system ensures that the future trajectory of the aircraft is always safe and uses projected distance maximization to ensure collision-free flight. The system performs a minimally invasive modification to the plans made by the planning/inference engine to ensure that these plans will not violate safety invariants. These safety invariants focus on avoiding collisions with other agents and risky maneuvers. The system assumes linear velocity vectors for other agents while interpolating for the future. The system is also designed with runtime performance objectives in mind and ensures safe behavior on the fly.

\subsection{Automated speech recognition and production}

Establishing clear communication between a human operator/pilot and an AI system in our target problem domain is critical. The challenge of understanding and decoding aviation-specific terminology that is different from everyday speech constructions is specific to the aviation domain. Other challenges, such as radio background noise, incomplete instructions, and radio phraseology, also need to be addressed.

We propose developing a bi-directional communication mechanism for human-AI collaboration, focusing on clarity instead of naturalness to accomplish an acceptable performance level to produce a language covering the controlled vocabulary used in airspace operations. We employed learning approaches for the AI system to understand complex concepts, e.g., learning from demonstrations, and, to language understanding, developed a language generation system, which can summarize visually perceived information~\cite{Junjiao2019}. This language generation capability can also clarify potential ambiguity when receiving commands from a human operator/pilot. To build a language representation model customized for aviation, we will leverage a large corpus of aviation-relevant documents such as manuals to train a model, and, to use such unlabeled data, we employed a transformer-based BERT~\cite{devlin2018bert} variant to pre-train from the unlabeled text.

\subsection{High-fidelity Simulators}

\begin{figure*}[thpb]
\centering
\includegraphics[width=\textwidth]{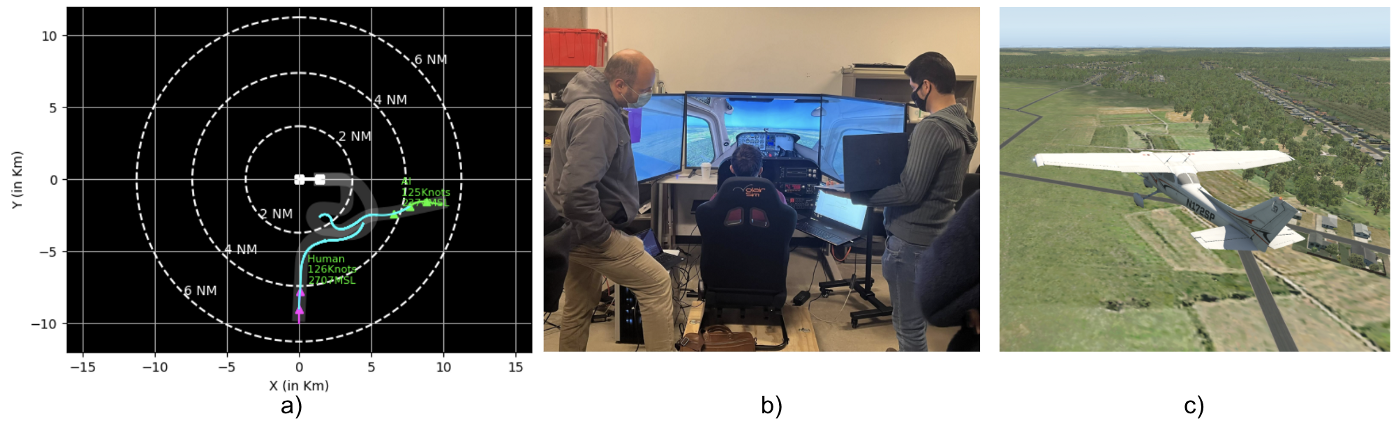}
\caption{The figure shows the high-fidelity simulator setup that enables Human-AI interaction. Figure a) shows a top-down view with one human agent (magenta) interacting and one AI agent (lime) while trying to land on the same runway. The most likely branch of the MCTS forward propagation tree for both agents is shown in cyan. White lines show the reference trajectories. Figure b) shows the physical simulator setup with an immersive environment for the human pilot. Figure c) shows a screengrab of the visual rendering of the simulator using the X-Plane 11 flight simulator backend.}
\label{fig:4}
\end{figure*}

For safety-critical domains, accurate high-fidelity simulators are required to test algorithms before real-world testing. Given the lack of simulators in the public domain, we designed our simulator christened XPlaneROS. With X-Plane 11 as the high-fidelity flight simulator and ROSplane~\cite{ellingson2017rosplane} as the autopilot, we obtain realistic aircraft models and visuals of similar aircraft (see Fig.~\ref{fig:4}). The X-Plane Connect Toolbox interfaces between X-Plane and Robot Operating System (ROS) topics. Based on the high-level commands input by the planner, ROSPlane generates the actuator commands for ailerons, rudder, elevator, and throttle, which are, in turn, sent to X-Plane through XPlaneConnect. ROSplane uses a cascaded control structure and can follow waypoints with Dubin’s Paths. XPlaneROS provides additional capabilities to follow a select set of motion primitives. There have also been extensions to ROSplane, like employing a proper takeoff, additional control loops for vertical velocity rates, and a rudimentary autonomous landing sequence.% Check the XPlaneROS architecture in Fig.~\ref{fig:5}

% \begin{figure}[hhh]
% \centering
% \includegraphics[width=0.48\textwidth]{fig5.png}
% \caption{A pictorial overview of the XPlaneROS.}
% \label{fig:5}
% \end{figure}

\section{Interactive Demonstration Summary}
\label{sec:3}
For the interactive demonstration, we propose a scenario with one AI pilot and one human-pilot landing on the same runway in a non-towered airfield with runways 08 and 26. The goal of the AI system is to execute a low approach over Runway 26, use a natural language process algorithm to understand the human intent, and coordinate with the human pilot to achieve the joint objective (see Fig.~\ref{fig:6}). The interactive demonstration has the following stages:

\begin{figure}[thpb]
\centering
\includegraphics[width=0.485\textwidth]{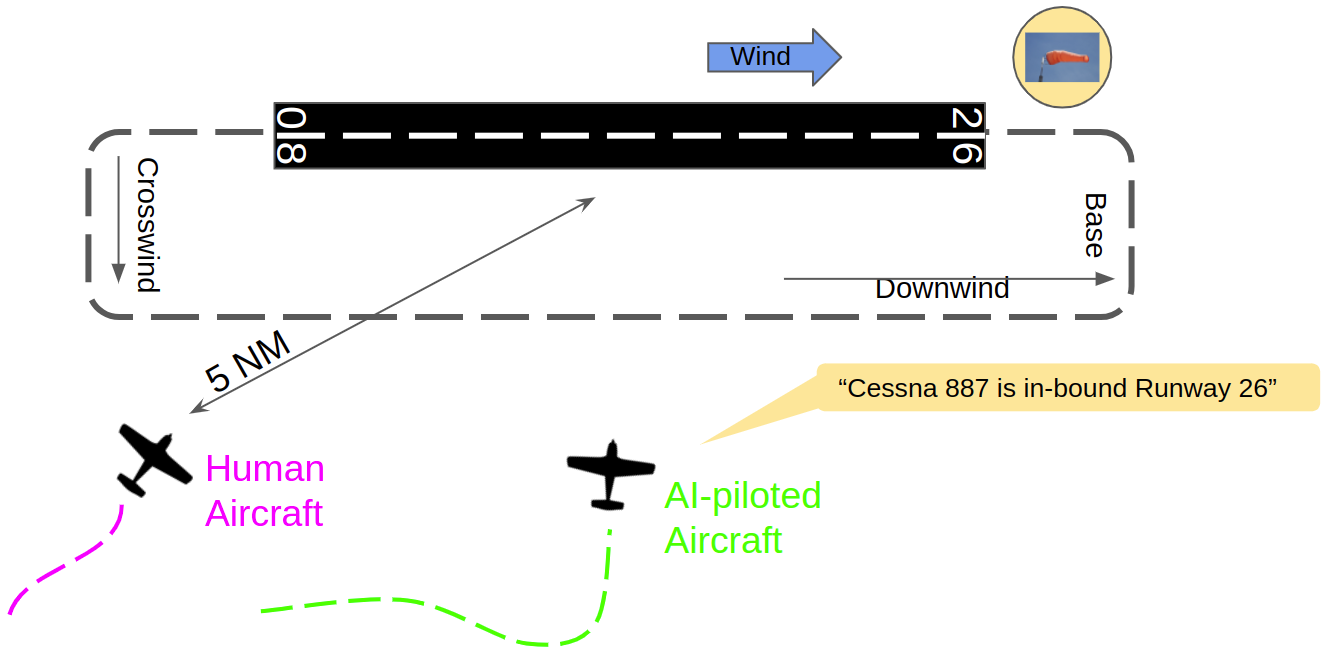}
\caption{Scenario depiction of the interactive demonstration.}
\label{fig:6}
\end{figure}

\begin{enumerate}
\item The human pilot is erroneously heading to land on runway 08 and informs their intention on a broadcast message.

\item The AI pilot generates a safe plan by predicting the human pilot's intention from the language by listening to the human pilot's broadcast message. 

\item The AI pilot correctly calculates that the preferred runway is 26 and generates speech to broadcast its goal of landing on runway 26 to the human pilot.

\item The human pilot decided to change to runway 26 and inform this new intention on a new broadcast message.

\item The AI pilot generates the new plan changing to runway 26. Once more, the AI pilot generates speech to broadcast its new goal to the human pilot.
 
\item Both the AI and the human pilots, knowing the position of each other, coordinate the landing on runway 26.
    
\end{enumerate}

A demonstration of vision-only long-range aircraft detection is also showcased using real-world data. 

\section{Conclusions}
\label{sec:4}

We believe the current understanding of integrating unmanned aircraft within the NAS needs a more human-pilot-centered approach. Pilots and aircraft already operating in the national airspace are essential stakeholders in the larger discourse.

Technological developments enabling the safe and seamless operation of manned and unmanned aircraft in shared airspace need an understanding of the rules and conventions already in place. The NAS is a dynamic environment with in-built flexibility and protocols to handle traffic volumes and emergencies. Our core understanding is that rather than having the current NAS adapt to changing autonomy needs, we need to move towards identifying technological requirements that enable unmanned systems to operate with humans collaboratively. We hope to build true AI pilots indistinguishable from human pilots to allow seamless integration within the current NAS.

\addtolength{\textheight}{-12cm}   % This command serves to balance the column lengths

%%%%%%%%%%%%%%%%%%%%%%%%%%%%%%%%%%%%%%%%%%%%%%%%%%%%%%%%%%%%%%%%%%%%%%%%%%%%%%%%

%%%%%%%%%%%%%%%%%%%%%%%%%%%%%%%%%%%%%%%%%%%%%%%%%%%%%%%%%%%%%%%%%%%%%%%%%%%%%%%%



\section*{ACKNOWLEDGMENT}

This work is supported by the Army Futures Command Artificial Intelligence Integration Center (AI2C).

%%%%%%%%%%%%%%%%%%%%%%%%%%%%%%%%%%%%%%%%%%%%%%%%%%%%%%%%%%%%%%%%%%%%%%%%%%%%%%%%

\bibliographystyle{ieeetran}  %%%% https://www.overleaf.com/learn/latex/bibtex_bibliography_styles

\bibliography{ref.bib}

\begin{thebibliography}{1}
\providecommand{\url}[1]{#1}
\csname url@rmstyle\endcsname
\providecommand{\newblock}{\relax}
\providecommand{\bibinfo}[2]{#2}
\providecommand\BIBentrySTDinterwordspacing{\spaceskip=0pt\relax}
\providecommand\BIBentryALTinterwordstretchfactor{4}
\providecommand\BIBentryALTinterwordspacing{\spaceskip=\fontdimen2\font plus
\BIBentryALTinterwordstretchfactor\fontdimen3\font minus
  \fontdimen4\font\relax}
\providecommand\BIBforeignlanguage[2]{{%
\expandafter\ifx\csname l@#1\endcsname\relax
\typeout{** WARNING: IEEEtran.bst: No hyphenation pattern has been}%
\typeout{** loaded for the language `#1'. Using the pattern for}%
\typeout{** the default language instead.}%
\else
\language=\csname l@#1\endcsname
\fi
#2}}

\bibitem{aweiss2018}
\BIBentryALTinterwordspacing
A.~S. Aweiss, B.~D. Owens, J.~Rios, J.~R. Homola, and C.~P. Mohlenbrink,
  \emph{Unmanned Aircraft Systems (UAS) Traffic Management (UTM) National
  Campaign II}.\hskip 1em plus 0.5em minus 0.4em\relax Aerospace Research
  Central, 2018. [Online]. Available:
  \url{https://arc.aiaa.org/doi/abs/10.2514/6.2018-1727}
\BIBentrySTDinterwordspacing

\bibitem{mueller2017enabling}
E.~R. Mueller, P.~H. Kopardekar, and K.~H. Goodrich, ``Enabling airspace
  integration for high-density on-demand mobility operations,'' in \emph{17th
  AIAA Aviation Technology, Integration, and Operations Conference}, 2017, p.
  3086.

\bibitem{lin2014microsoft}
T.-Y. Lin, M.~Maire, S.~Belongie, J.~Hays, P.~Perona, D.~Ramanan,
  P.~Doll{\'a}r, and C.~L. Zitnick, ``Microsoft coco: Common objects in
  context,'' in \emph{European conference on computer vision}.\hskip 1em plus
  0.5em minus 0.4em\relax Springer, 2014, pp. 740--755.

\bibitem{patrikar2021trajair}
\BIBentryALTinterwordspacing
J.~Patrikar, B.~Moon, J.~Oh, and S.~Scherer, ``Predicting like a pilot: Dataset
  and method to predict socially-aware aircraft trajectories in non-towered
  terminal airspace,'' 2021. [Online]. Available:
  \url{https://arxiv.org/abs/2109.15158}
\BIBentrySTDinterwordspacing

\bibitem{Junjiao2019}
\BIBentryALTinterwordspacing
J.~Tian and J.~Oh, ``Image captioning with compositional neural module
  networks,'' in \emph{Proceedings of the Twenty-Eighth International Joint
  Conference on Artificial Intelligence, {IJCAI-19}}.\hskip 1em plus 0.5em
  minus 0.4em\relax International Joint Conferences on Artificial Intelligence
  Organization, 7 2019, pp. 3576--3584. [Online]. Available:
  \url{https://doi.org/10.24963/ijcai.2019/496}
\BIBentrySTDinterwordspacing

\bibitem{devlin2018bert}
J.~Devlin, M.-W. Chang, K.~Lee, and K.~Toutanova, ``Bert: Pre-training of deep
  bidirectional transformers for language understanding,'' \emph{arXiv preprint
  arXiv:1810.04805}, 2018.

\bibitem{ellingson2017rosplane}
G.~Ellingson and T.~McLain, ``Rosplane: Fixed-wing autopilot for education and
  research,'' in \emph{Unmanned Aircraft Systems (ICUAS), 2017 International
  Conference on}.\hskip 1em plus 0.5em minus 0.4em\relax IEEE, 2017.

\end{thebibliography}

\end{document}